# Accurate 3D Prediction of Missing Teeth in Diverse Patterns for Precise Dental Implant Planning

Lei Ma[a,1], Peng Xue[a,b,1], Yuning Gu[a], Yue Zhao[c], Min Zhu[d], Zhongxiang Ding[e], Dinggang Shen[a,f,g,*]

[a]*School of Biomedical Engineering, ShanghaiTech University, Shanghai 201210, China.*
[b]*School of Mechanical, Electrical and Information Engineering, Shandong University, Weihai, 264209, China*
[c]*School of Communication and Information Engineering, Chongqing University of Posts and Telecommunications, Nan'an District, Chongqing, 400065, China.*
[d]*Shanghai Ninth People's Hospital, Shanghai Jiao Tong University, Huangpu District, Shanghai, 200011, China.*
[e]*Department of Radiology, Hangzhou First People's Hospital, Zhejiang University, Hangzhou, 310006, China.*
[f]*Shanghai United Imaging Intelligence Co., Ltd., Shanghai 200230, China, and Shanghai Clinical Research and Trial Center, Shanghai 201210, China.*
[g]*Shanghai Clinical Research and Trial Center, Shanghai, 201210, China.*

**Abstract**

In recent years, the demand for dental implants has surged, driven by their high success rates and esthetic advantages. However, accurate prediction of missing teeth for precise digital implant planning remains a challenge due to the intricate nature of dental structures and the variability in tooth loss patterns. This study presents a novel framework for accurate prediction of missing teeth in different patterns, facilitating digital implant planning. The proposed framework begins by estimating point-to-point correspondence among a dataset of dental mesh models reconstructed from CBCT images of healthy subjects. Subsequently, tooth dictionaries are constructed for each tooth type, encoding their position and shape information based on the established point-to-point correspondence. To predict missing teeth in a given dental mesh model, sparse coefficients are learned by sparsely representing adjacent teeth of the missing teeth using the corresponding tooth dictionaries. These coefficients are then applied to the dictionaries of the missing teeth to generate accurate predictions of their positions and shapes. The evaluation results on real subjects shows that our proposed framework achieves an average prediction error of 1.04mm for predictions of single missing tooth and an average prediction error of 1.33mm for the prediction of 14 missing teeth, which demonstrates its capability of accurately predicting missing teeth in various patterns. By accurately predicting missing teeth, dental professionals can improve the planning and placement of dental implants, leading to better esthetic and functional outcomes for patients undergoing dental implant procedures.

## 1. Introduction

Dental implantation has gained significant popularity as a reliable solution for restoring missing teeth due to its high success rates and esthetic benefits [1, 2]. It involves surgically placing an artificial tooth root into the jawbone, which provides a stable foundation for a replacement tooth [3]. The detailed workflow of dental implantation includes the following steps: an initial consultation and examination, pre-operative imaging such as taking CBCT scan, dental implant planning, and final restoration of the implant with a prosthetic tooth [4]. Notably, den-

---

tal implant planning plays a crucial role in ensuring the accuracy and success of the implant placement [5]. With the advancements in dental imaging and computer-aided design (CAD) technologies, dental implant planning has transitioned into a digitalized process, offering improved precision, efficiency, and predictable outcomes [6]. The process of digital implant planning typically involves acquiring high-quality 3D imaging, processing the images in specialized software, virtual tooth positioning based on the patient's anatomy and treatment goals, and creating virtual mock-ups to visualize the final outcome [7]. Moreover, the digital implant planning system empowers dentists to optimize the position and orientation of the implant in the virtual environment, ensuring the best possible outcome for the patient.

The key step in digital implant planning is the virtual tooth positioning, which involves the use of computer-aided design (CAD) software to virtually plan and position dental implants [8]. This step enables dentists to determine the optimal location, angle, and depth for implant placement, resulting in functional and natural-looking tooth replacements [9]. In current clinical practice, dentists typically manually perform virtual tooth positioning in digital implant planning using specialized software that allows them to manipulate virtual tooth models based on patient's dental scans [10, 11]. However, this manual process heavily relies on the dentist's experience and expertise in digital implant planning, often involving trial and error to achieve the desired virtual tooth positioning based on each patient's unique oral anatomy. Consequently, manual virtual tooth positioning can be time-consuming and labor-intensive, requiring careful analysis and manipulation of virtual tooth models to achieve optimal position and aesthetics. This can increase overall treatment planning time and demand additional effort and resources.

Accurate prediction of positions and shapes of missing teeth can provide references for virtual tooth positioning, enabling dentists to achieve optimal outcomes with greater efficiency [12]. However, accurately predicting missing teeth remains a challenge due to the complexity of dental structures and the variability of tooth loss patterns. Specifically, the dental structure is inherently intricate, with each tooth possessing a unique shape and position that can vary among individuals [13]. Additionally, tooth loss patterns can vary widely, ranging from single missing teeth to multiple missing teeth with different locations and orientations [14]. Previously, few efforts have been made to address the challenge of accurately predicting the position and shape of missing teeth. Part et al. proposed a deep learning-based method to detect regions of missing teeth in panoramic radiographic images for assisting dental implant placement planning [15]. Bayrakdar et al. developed a framework to determine localization of missing teeth based on their neighboring teeth locations and surrounding jaw structures [12]. However, these methods are unable to directly predict positions and 3D shapes of missing teeth, thus limiting their applications in dental implant planning. Therefore, there is a critical need for a more reliable and accurate method for predicting missing teeth placements and shapes in digital implant planning.

In this study, we introduce a novel framework designed to accurately predict positions and shapes of missing teeth in different patterns, thereby facilitating precision digital implant planning. Our framework commences by estimating the point-to-point correspondence among a dataset of dental mesh models reconstructed from cone-beam computed tomography (CBCT) images of healthy subjects. Subsequently, tooth dictionaries are constructed for each tooth type, encoding tooth position and shape information based on the established point-to-point correspondence. By sparsely representing the adjacent teeth using their tooth dictionaries, a set of sparse coefficients is learned for predicting missing teeth. These coefficients are then applied to the dictionaries of the missing teeth to generate accurate predictions of their positions and shapes. The framework's ability to predict missing teeth based on their adjacent teeth enables accurate predictions across various types of tooth loss. This design confers flexibility and adaptability to different clinical scenarios, rendering it a valuable tool for dental practitioners in addressing diverse cases of tooth loss during implant planning procedures. The contributions of this paper can be summarized as follows: (1) We propose a novel framework that accurately predicts missing teeth for precision digital implant planning using a CBCT image dataset of healthy subjects. (2) The proposed framework leverages the sparse representation of the remaining adjacent teeth to accurately predict missing teeth in various patterns. (3) The proposed framework is evaluated on real patients, adding practicality and applicability to the findings. The results validate its potential as a valuable tool for dental practitioners in improving

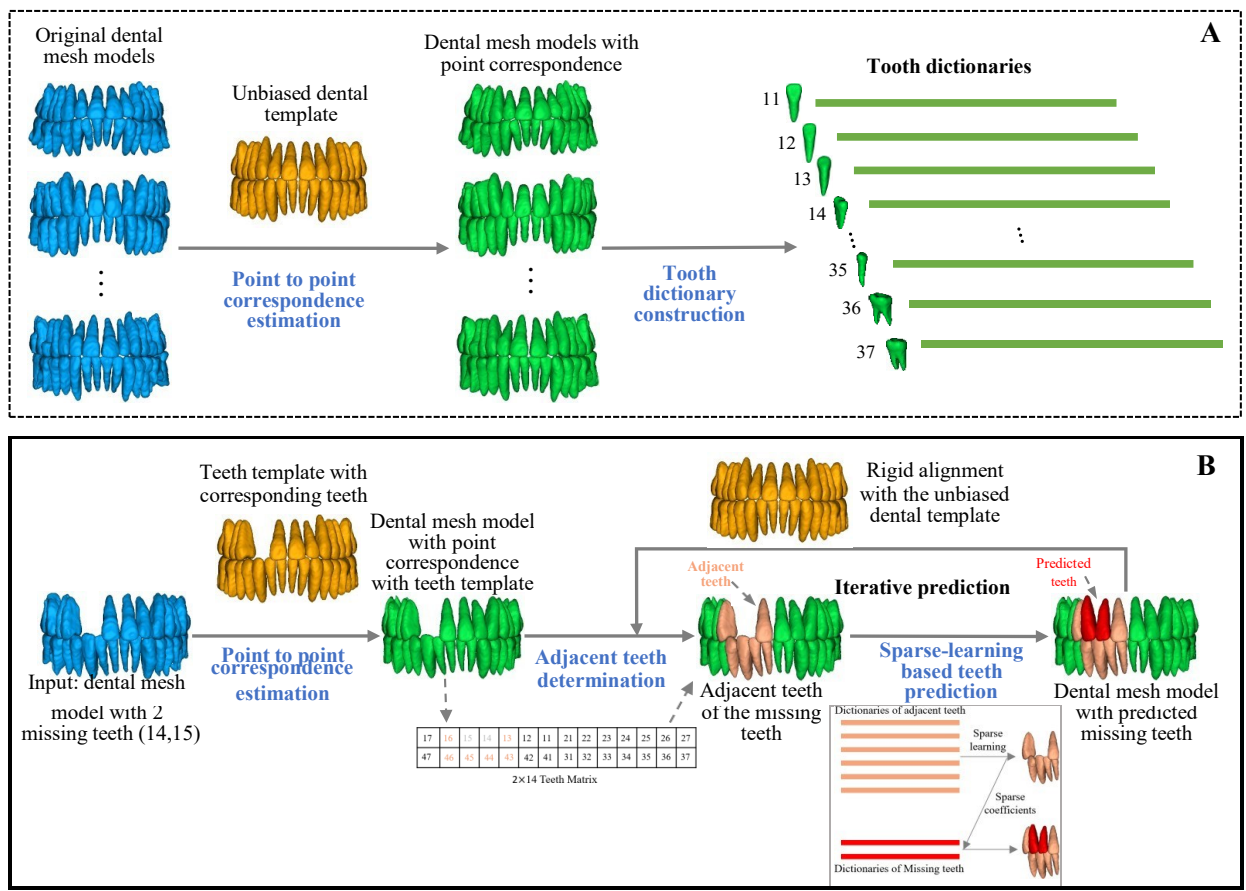

Figure 1: (A) The process of tooth dictionary construction. (B) The proposed framework for predicting missing teeth in a given dental mesh model.

digital implant planning accuracy and efficiency.

The rest of the paper is organized as follows. The studied data and proposed method are described in Section 2. The evaluation results of our proposed framework on missing teeth prediction in various patterns are presented in Section 3. The paper is finally discussed and concluded in Section 4.

## 2. Methods

The proposed framework for accurate prediction of missing teeth is illustrated in Fig. 1. To begin, we estimate the point-to-point correspondence among dental mesh models from different subjects in our dataset. This is achieved by leveraging an unbiased dental template that serves as a reference [13]. Next, we construct a dictionary for each type of tooth present in the dataset. These dictionaries are created to encode the positions and shapes of the respective teeth. They capture the characteristic features and variations observed among the teeth in the dataset. To predict the missing teeth in a given dental mesh model, we employ a sparse learning-based method to represent the adjacent teeth using their corresponding tooth dictionaries, resulting in a group of sparse coefficients. Finally, we apply the obtained coefficients to the dictionary specific to the missing tooth types, generating accurate predictions for the missing teeth.

### 2.1. Data and pre-processing

In this study, cone-beam computed tomography (CBCT) images of 133 healthy subjects were selected

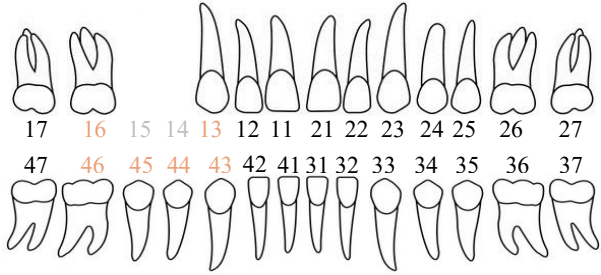

An example of dental model with missing teeth 14 and 15

| 17 | 16 | 15 | 14 | 13 | 12 | 11 | 21 | 22 | 23 | 24 | 25 | 26 | 27 |
|----|----|----|----|----|----|----|----|----|----|----|----|----|----|
| 47 | 46 | 45 | 44 | 43 | 42 | 41 | 31 | 32 | 33 | 34 | 35 | 36 | 37 |

2×14 Teeth Matrix

Figure 2: An example of dental mesh model with missing teeth 14 and 15, and its corresponding teeth matrix. In our implementation, *t* is set to 1. According to the proposed approach, the teeth colorized in the matrix are determined as adjacent teeth of the missing teeth.

from our digital archive. These images were acquired using a Planmeca dental CBCT scanner. Subjects with severe malocclusion or missing teeth were excluded during the data selection process. Additionally, the third molars were not considered in this study. The selected subjects were of Chinese ethnicity and ranged in age from 18 to 60 years.

Data preprocessing was performed to reconstruct dental mesh models from these CBCT images. First, a fully-automatic deep learning-based tooth segmentation method [16] was employed to segment teeth from the CBCT images. Next, a dental atlas-based teeth labeling method [13] was applied to label the segmented teeth. Finally, dental mesh models were reconstructed from the labeled teeth segmentation images. This reconstruction step allowed for the representation of the teeth in a three-dimensional mesh format, which was essential for subsequent analyses and predictions within our framework.

*2.2. Tooth dictionary construction*

To construct a dictionary for each type of tooth, it is essential to estimate point-to-point correspondences across the teeth within the type. This is achieved by leveraging an unbiased dental template developed in [13]. The unbiased dental template provides a population-average representation of typical tooth shapes and locations while maintaining high cross-individual validity within the population, thus enhancing the accuracy of point correspondence estimation [17]. The estimation of point correspondence involves the following steps:

1) Rigid Alignment: Initially, the dental mesh models derived from CBCT images are rigidly aligned to the unbiased dental template. This alignment ensures consistent positioning and orientation across the dental mesh models.

2) Non-rigid Deformation: Next, each tooth in the dental template is non-rigidly deformed to match its corresponding tooth in the aligned dental mesh models. This deformation is accomplished using the coherent point drift (CPD) algorithm [18]. The CPD algorithm effectively captures shape variations and deformations specific to each tooth, enabling accurate point correspondence.

3) Nearest Point Search: For each point in the deformed tooth shape, the nearest point in its target tooth shape is identified. This process establishes a tooth shape with strict point-to-point correspondence to its corresponding tooth template.

Through these steps, we can accurately estimate strict point-to-point correspondence for each type of tooth in the dental mesh models.

In the human adult dentition, excluding the third molars, there are typically 28 teeth. In this study, we treat each tooth as an individual type, resulting in 28 distinct tooth types. For each tooth type, we construct a dictionary based on the previously estimated point-to-point correspondence. Specifically, we transform the coordinates ($x$, $y$, $z$) of all the corresponding points of the $i$-th tooth type in the studied dataset into a $3T_i \times N$ matrix. Here, $T_i$ represents the number of points in the $i$-th tooth type, and $N$ indicates the number of subjects used for dictionary construction. The resulting matrix, denoted as $D_i$, represents the dictionary for the $i$-th tooth type. It is a $3T_i \times N$ matrix, where each column corresponds to the coordinates ($x$, $y$, $z$) of the corresponding points for one subject, and every three rows represent the same corresponding point across all subjects in the dataset. By following this process, we construct a total of 28 tooth dictionaries, denoted as ($D_1$, $D_2$, ..., $D_{28}$), encompassing all tooth types present in the dental mesh models.

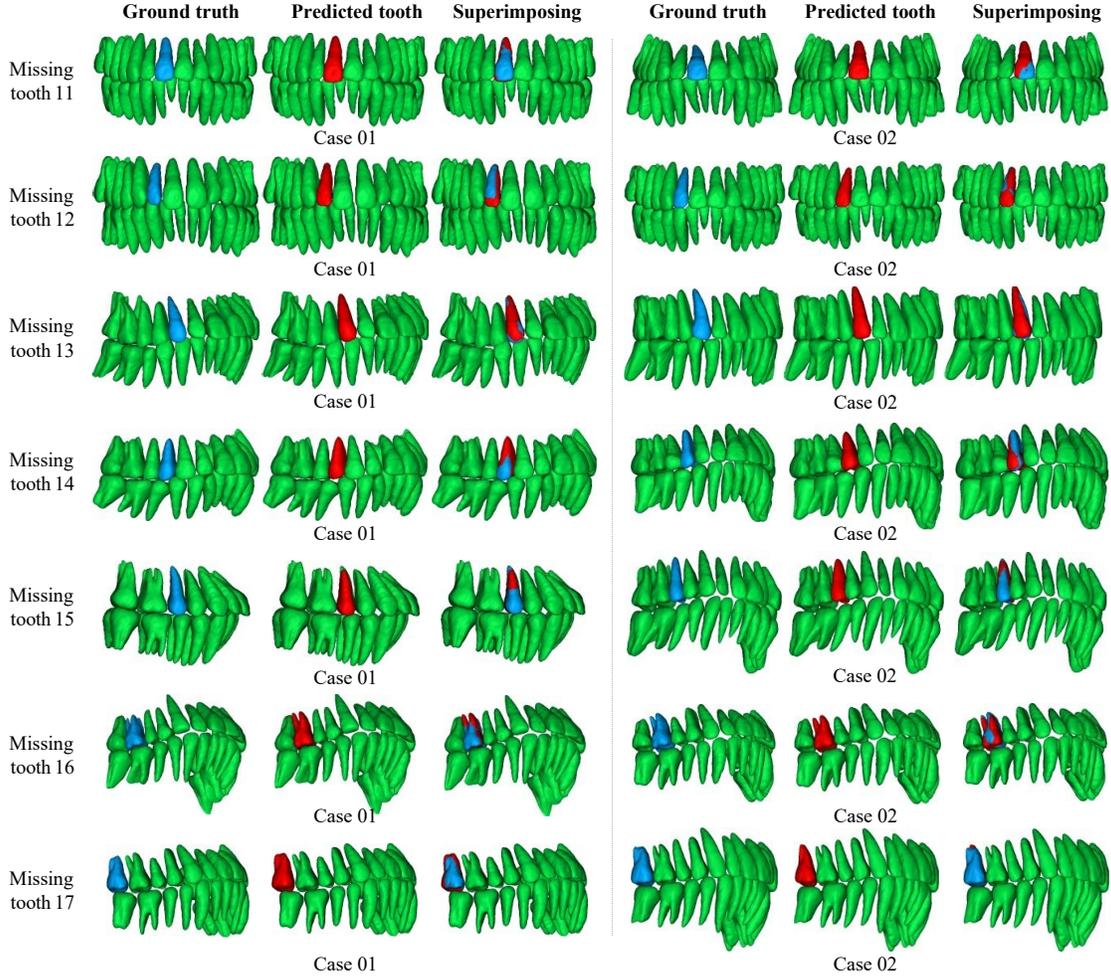

Figure 3: Visual comparisons between the predicted teeth and their corresponding ground truths in the different patterns of single missing tooth. The blue and red teeth represent the ground-truth teeth and predicted teeth, respectively.

### 2.3. Adjacent teeth determination

Given a dental mesh model with missing teeth $(M_1, M_2, ..., M_k)$, we predict the missing teeth by relying on the positions and shapes of their adjacent teeth $(A_1, A_2, ..., A_h)$. Here, $k$ represents the number of missing teeth, and $h$ denotes the number of adjacent teeth. In order to handle various patterns of missing teeth, we have developed an effective approach to determine the adjacent teeth used for prediction. Specifically, we arrange the 28 human teeth in a 2×14 teeth matrix, as illustrated in Fig. 2. Based on the positions of the missing teeth in the teeth matrix, we can derive their column index range, denoted as $(CR_{min}, CR_{max})$. This range provides information on the relative positions of the missing teeth within the teeth matrix. Consequently, we determine the adjacent teeth for prediction by considering the existing teeth within the column range of $(CR_{min} - t, CR_{max} + t)$, where $t \geq 1$. These adjacent teeth play a crucial role in accurately predicting the missing teeth. In cases where $CR_{min} - t \leq 0$, indicating that the required adjacent teeth extend beyond the left border of the teeth matrix (the second molars 17 and 47), we determine the adjacent teeth between the column range of $(CR_{min} - t + 14, 14)$ and $(1, CR_{max} + t)$. This approach leverages the teeth on the opposite side of the teeth matrix ($CR_{min}$ -t+14, 14) to improve the prediction accuracy, particularly for missing molars. Similarly, if $CR_{max} + t > 14$, we determine the adjacent teeth as the teeth between the column range of ($CR_{min}$-t,14) and (1,$CR_{max}$ +t-14). This accounts for cases where the adjacent teeth extend beyond the right border of the teeth matrix. By considering the appropriate adjacent teeth based on the column index range, our framework can accurately predict the missing teeth in different patterns.

### 2.4. Iterative missing teeth prediction

To accurately predict the positions and shapes of missing teeth from their adjacent teeth, we propose an iterative sparse learning-based method. The method consists of the following steps:

1) Initially, we rigidly align the dental mesh model with the unbiased teeth template relying on the remaining teeth. Note that the teeth corresponding to the missing teeth are also removed from the teeth template.

2) Next, we estimate the point-to-point correspondence between each tooth in the aligned dental mesh model and its corresponding tooth in the dental template.

3) Further, we sparsely represent the adjacent teeth $A_{adj} = (A_1; A_2; ...; A_h)$ using the combined tooth dictionaries of the adjacent teeth $D_{adj} = (D_{A_1}; D_{A_2}; ...; D_{A_h})$ by solving the following $l1$-minimization problem [19, 20]:

$$\hat{C} = \text{argmin} \|C\|_1 \text{ subject to} \|D_{adj}C - A_{adj}\|_2 \leq \varepsilon, \quad (1)$$

where $\hat{C}$ denotes the coefficients estimated in the sparse representation. $\varepsilon$ represents an error tolerance.

4) We predict the shape of the missing tooth $\hat{M}_i$ using the following equation:

$$\hat{M}_i = D_{M_i}\hat{C}, \quad (2)$$

where $D_{M_i}$ is the dictionary of the missing tooth $M_i$.

5) To refine the tooth prediction, we rigidly align the dental mesh model merged with the teeth prediction to the unbiased teeth template. Then, we repeat steps 3 and 4 to further improve the accuracy of the tooth prediction.

These steps (i.e., step 3, step 4, and step 5) are performed iteratively until the prediction error of the missing teeth converges. This iterative process helps gradually reduce the prediction error caused by the registration error of the initial rigid alignment between the dental mesh model with missing teeth and the unbiased teeth template.

## 3. Experiments and Results

### 3.1. Experimental Setup

We conducted experiments to evaluate the effectiveness of the proposed method for predicting missing teeth. From the studied dataset, we randomly selected 100 subjects to construct the tooth dictionaries, while the remaining 33 subjects were used for testing. In this implementation, we set $t$ to 1 for determining the adjacent teeth. We repeated the iterative teeth prediction process three times to refine the tooth predictions. To evaluate the performance of the method, we conducted both qualitative and quantitative assessments on the testing results. For the qualitative evaluation, we visually compared the predicted teeth with their corresponding ground-truth teeth. In addition, we performed a quantitative evaluation by estimating the teeth prediction error. The prediction error was computed by averaging the point distances between the points in the predicted teeth and their corresponding points in the ground-truth teeth. Furthermore, we estimated the shape error of the predicted teeth by first rigidly aligning the predicted teeth to their ground truths and then measuring the point cloud distance between them [21]. By conducting these evaluations, we were able to assess both visual quality and quantitative accuracy of the predicted teeth compared to their ground truth.

### 3.2. Results of Single Missing Tooth Prediction

We first performed experiments to evaluate the effectiveness of our proposed method for single missing tooth prediction. For each experiment, we removed one tooth from each dental mesh model in the testing set and utilized our method to predict the removed (missing) tooth. This resulted in a total of 28 experiments to predict all types of teeth.

Figure 3 showcases visual comparisons between the predicted teeth (red) and their corresponding ground truth (blue). We selected seven types of teeth, specifically from

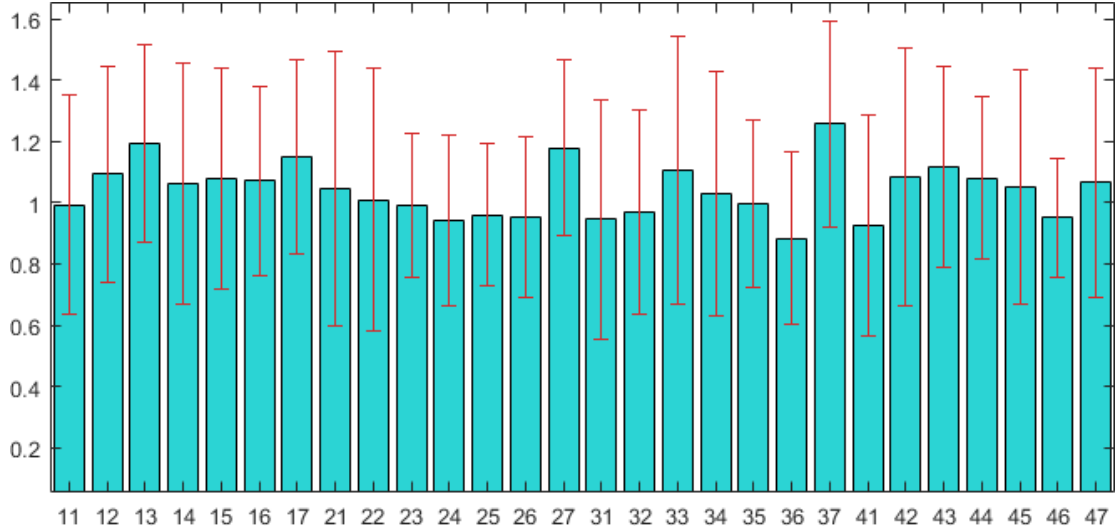
Figure 4: The means and standard deviations of the prediction error of all teeth in single missing teeth prediction experiments.

Table 1: Average shape errors of the predicted teeth. The odd rows represent tooth numbers, and the even rows indicate the estimated shape errors (mm) of the predicted teeth.

| 11 | 12 | 13 | 14 | 15 | 16 | 17 | 21 | 22 | 23 | 24 | 25 | 26 | 27 |
|---|---|---|---|---|---|---|---|---|---|---|---|---|---|
| 0.80 | 0.75 | 0.81 | 0.78 | 0.77 | 0.75 | 0.91 | 0.74 | 0.73 | 0.77 | 0.72 | 0.70 | 0.69 | 0.94 |
| 31 | 32 | 33 | 34 | 35 | 36 | 37 | 41 | 42 | 43 | 44 | 45 | 46 | 47 |
| 0.63 | 0.71 | 0.77 | 0.77 | 0.74 | 0.65 | 0.93 | 0.63 | 0.72 | 0.82 | 0.78 | 0.74 | 0.72 | 0.79 |

teeth 11 to 17, for visual comparison and randomly chose two different cases for each tooth type. To facilitate clear comparison, we also superimposed the predicted teeth with their ground truth and displayed them alongside the remaining teeth (green) in the tooth model.

Figure 4 presents the averaged prediction errors and standard deviations of the predicted teeth across different types. The average prediction errors ranged from a minimum of 0.88mm (tooth 36) to a maximum of 1.26mm (tooth 37). The mean prediction error across all teeth was 1.04mm. Table 1 reports the estimated shape errors for different types of teeth. The maximum shape error was 0.94mm (tooth 27), while the minimum was 0.63mm (tooth 31 and 41). These results demonstrate accurate prediction capabilities of our proposed method for various types of missing teeth. Furthermore, it is noteworthy that the prediction errors of second molars (tooth 17, tooth 27, tooth 37, and tooth 47) were slightly larger compared to other tooth types. This can be attributed to the fact that second molars have adjacent teeth on only one side, unlike other teeth (see Figure 2).

### 3.3. Results of Multiple Missing Teeth Prediction

We further conducted experiments to evaluate the performance of our proposed method on multiple missing teeth prediction. We simulated various scenarios of multiple missing teeth using the testing dataset. Specifically, we simulated missing teeth in the same tooth row, ranging from 2 missing teeth to 7 missing teeth. Additionally, we simulated missing teeth in both the upper and lower tooth rows.

Table 2 presents the prediction errors of multiple missing teeth in the same tooth row, while Table 3 shows the prediction errors of multiple missing teeth in different tooth rows. These results demonstrate that our method can accurately predict multiple missing teeth, even up to 14 missing teeth, in different patterns. Specifically, the average prediction error of the prediction of the 14 missing teeth was 1.33mm, demonstrating the effectiveness of our framework in handling a large number of missing teeth.

Table 2: Average shape errors of the predicted teeth. The odd rows represent tooth numbers of multiple missing teeth in the same tooth row, and the even rows indicate the prediction errors (mm) of the missing teeth.

|  |  |  |  |  | 12 | 11 |
|---|---|---|---|---|---|---|
|  |  |  |  |  | 1.07 | 1.09 |
|  |  |  |  | 13 | 12 | 11 |
|  |  |  |  | 1.31 | 1.11 | 1.18 |
|  |  |  | 14 | 13 | 12 | 11 |
|  |  |  | 1.15 | 1.38 | 1.04 | 1.10 |
|  |  | 15 | 14 | 13 | 12 | 11 |
|  |  | 1.17 | 1.25 | 1.43 | 1.12 | 1.12 |
|  | 16 | 15 | 14 | 13 | 12 | 11 |
|  | 1.19 | 1.16 | 1.22 | 1.34 | 1.12 | 1.15 |
| 17 | 16 | 15 | 14 | 13 | 12 | 11 |
| 1.20 | 1.16 | 1.19 | 1.21 | 1.31 | 1.14 | 1.16 |

Table 3: Average shape errors of the predicted teeth. The odd rows represent tooth numbers of multiple missing teeth in different tooth rows, and the even rows indicate the prediction errors (mm) of the missing teeth.

|  |  |  |  |  | 11 | 41 |  |  |  |  |  |  |
|---|---|---|---|---|---|---|---|---|---|---|---|---|
|  |  |  |  |  | 1.09 | 1.07 |  |  |  |  |  |  |
|  |  |  |  | 12 | 11 | 41 | 42 |  |  |  |  |  |
|  |  |  |  | 1.14 | 1.06 | 1.10 | 1.08 |  |  |  |  |  |
|  |  |  | 13 | 12 | 11 | 41 | 42 | 43 |  |  |  |  |
|  |  |  | 1.32 | 1.13 | 1.15 | 1.14 | 1.19 | 1.22 |  |  |  |  |
|  |  | 14 | 13 | 12 | 11 | 41 | 42 | 43 | 44 |  |  |  |
|  |  | 1.18 | 1.33 | 1.16 | 1.09 | 1.08 | 1.14 | 1.40 | 1.16 |  |  |  |
|  | 15 | 14 | 13 | 12 | 11 | 41 | 42 | 43 | 44 | 45 |  |  |
|  | 1.16 | 1.19 | 1.49 | 1.21 | 1.09 | 1.06 | 1.20 | 1.46 | 1.24 | 1.17 |  |  |
| 16 | 15 | 14 | 13 | 12 | 11 | 41 | 42 | 43 | 44 | 45 | 16 |  |
| 1.25 | 1.27 | 1.34 | 1.34 | 1.30 | 1.17 | 1.07 | 1.22 | 1.43 | 1.27 | 1.22 | 1.12 |  |
| 17 | 16 | 15 | 14 | 13 | 12 | 11 | 41 | 42 | 43 | 44 | 45 | 46 | 47 |
| 1.34 | 1.22 | 1.35 | 1.25 | 1.35 | 1.28 | 1.19 | 1.05 | 1.17 | 1.31 | 1.31 | 1.38 | 1.17 | 1.23 |

It is important to note, however, that the prediction errors tend to increase as the number of missing teeth increases. This observation aligns with our expectations, as the adjacent teeth used for prediction become further away from the missing teeth, leading to a weaker relationship between them and resulting in higher prediction errors. Figure 2 presents visual comparisons between the predicted multiple missing teeth and their corresponding ground-truth teeth. These visualizations provide further insights into the accuracy of our predictions in various patterns.

## 4. Discussions and Conclusions

In this paper, we present a novel framework for accurately predicting positions and shapes of missing teeth in dental implant planning. Our approach leverages the information of adjacent teeth to predict missing teeth, utilizing an iterative sparse learning-based method based on tooth dictionaries constructed from a dataset of normal subjects. Experimental results demonstrate high accuracy of our framework in predicting both the positions and shapes of missing teeth. Accurate prediction of tooth positions is crucial as it allows dentists to improve precision of tooth positioning during implant placement. Furthermore, accurate prediction of tooth shapes enables dentists to achieve aesthetically pleasing dental implants that can blend seamlessly with the patient's natural teeth. An important aspect of our proposed framework is its ability to address the challenges associated with multiple missing teeth in different patterns. By leveraging the adjacent teeth information, our framework is specifically designed to handle cases with multiple missing teeth, regardless of their arrangement or location. Experimental results on multiple teeth missing prediction further demonstrate the effectiveness of our proposed framework. Moreover, our framework extends its applicability beyond a single row of missing teeth, as it successfully predicts multiple missing teeth in both the upper and lower tooth rows. This capability enhances the versatility and practicality of our framework for a wide range of dental implant scenarios.

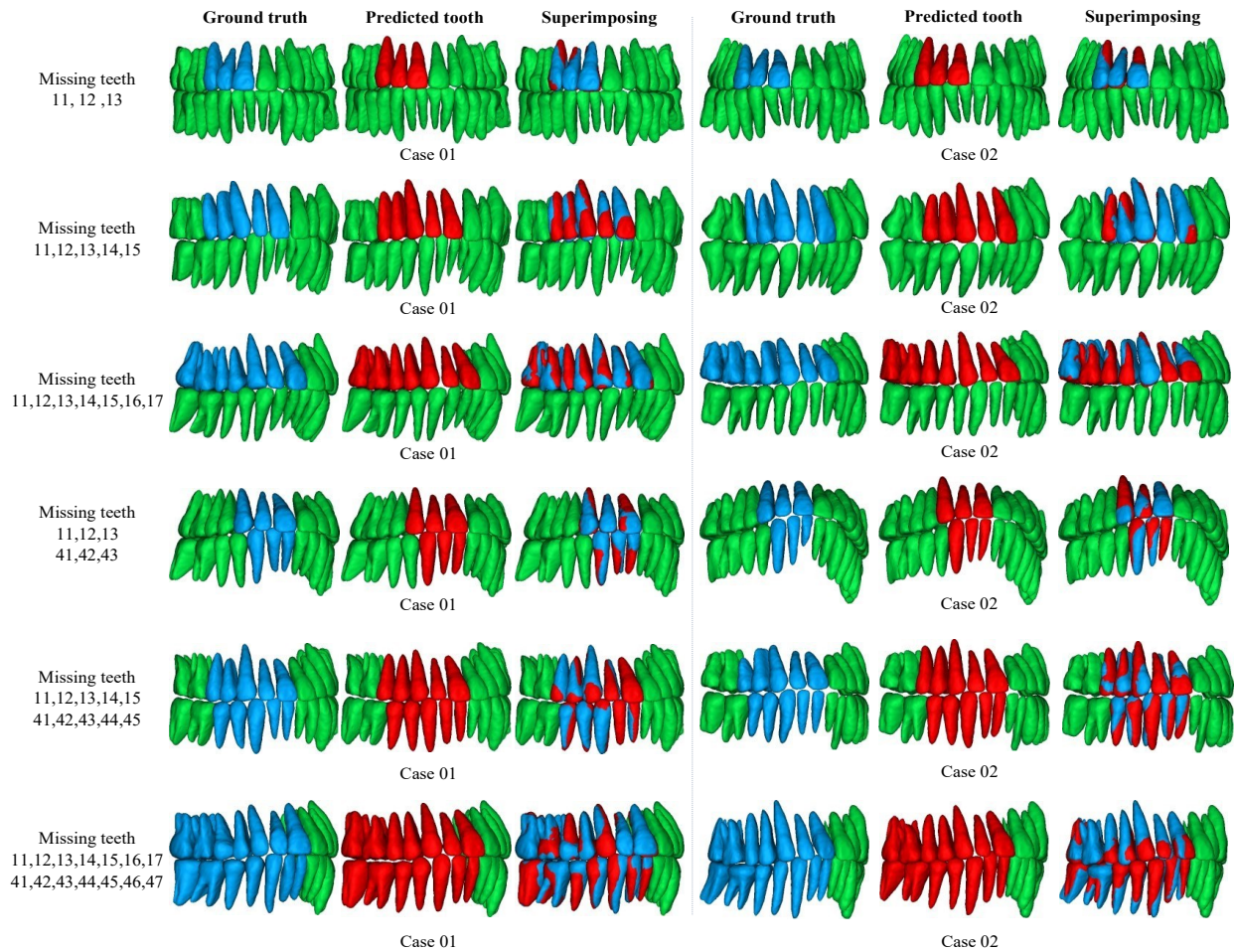

Figure 5: Visual comparisons between predicted missing teeth (red) and ground truth teeth (blue) in different patterns of multiple missing teeth.

The clinical significance of this study lies in its potential to improve dental treatment planning and patient care. By accurately predicting missing teeth in 3D dental mesh models, several benefits can be realized. First, accurate prediction of missing teeth allows dentists and orthodontists to better plan dental treatments, such as orthodontic procedures, dental implants, or prosthodontic interventions. It provides valuable information about the spatial arrangement and positions of the missing teeth, enabling more precise treatment strategies and outcomes. Second, missing teeth can have a significant impact on a patient's appearance and oral function. By predicting missing teeth accurately, dental professionals can develop comprehensive treatment plans that address both the aesthetic and functional aspects of the patient's dentition. This can lead to improved aesthetics, speech, chewing ability, and overall oral health. Third, traditional methods for predicting missing teeth often require extensive manual work and time-consuming procedures. The proposed 3D approach using sparse learning offers a more efficient and automated solution, potentially reducing the time and effort required for treatment planning. This can result in cost savings for both dental practitioners and patients. The last but not the least, the ability to predict missing teeth in different positions and with varying numbers allows for a more personalized approach to patient care. Treatment plans can be tailored to each individual's specific needs, taking into account their unique dental characteristics and requirements. This personalized approach can lead to improved patient satisfaction and outcomes.

While our proposed framework demonstrates promising results in predicting the positions and shapes of missing teeth, there are certain limitations that should be acknowledged. First, our framework heavily relies on the position and shape information provided by the adjacent teeth to predict the missing teeth. Therefore, inaccurate or missing information from the adjacent teeth may affect accuracy of the predictions. Second, the tooth dictionaries used in our framework are constructed from a dataset of normal subjects. This may introduce limitations when applying the framework to patients with unique dental characteristics that deviate from the dataset. These unique characteristics can include variations in tooth size, shape, alignment, or the presence of dental anomalies or abnormalities.

There are several ways for future research and development that can further enhance the accuracy and applicability of our approach. First, to improve generalizability, future work could involve expanding the dataset to include a more diverse range of dental characteristics, including variations in tooth morphology, size, and shape. This would allow for better representation of the population and enhance the framework's performance in predicting missing teeth for a wider range of patients. Second, our framework primarily focuses on the tooth structure and relies on the assumption that the surrounding bone and soft tissues are intact and suitable for dental implantation. External factors such as bone quality, gum health, and patient-specific conditions may influence final outcomes and should be considered in conjunction with our predictions in the future research. Third, future research could explore integration of patient-specific data, such as information from intraoral scans and patient medical history, for enhancing the accuracy and personalization of predictions. This could enable a more tailored approach to dental implant planning that considers individual patient's characteristics.

## Acknowledgments

This work was supported in part by National Natural Science Foundation of China (grant number 62131015), Science and Technology Commission of Shanghai Municipality (STCSM) (grant number 21010502600), and The Key R&D Program of Guangdong Province, China (grant number 2021B0101420006).